\documentclass[sigconf,letterpaper,9pt]{acmart}

\usepackage{booktabs} 
\usepackage{qtree} 
\usepackage{caption}
\usepackage{multirow}
\usepackage{subcaption}
\usepackage{changepage}

\setcopyright{rightsretained}

\makeatother

\begin{document}

\copyrightyear{2019}
\acmYear{2019}
\setcopyright{acmcopyright}
\acmConference[FAT* '19]{FAT* '19: Conference on Fairness, Accountability, and
Transparency}{January 29--31, 2019}{Atlanta, GA, USA}
\acmPrice{15.00}
\acmDOI{10.1145/3287560.3287600}
\acmISBN{978-1-4503-6125-5/19/01}

\title{50 Years of Test (Un)fairness: Lessons for Machine Learning}

\author{Ben Hutchinson and  Margaret Mitchell}
\email{{benhutch,mmitchellai}@google.com}

\begin{abstract}
Quantitative definitions of what is {\em unfair} and what is {\em fair} have been introduced in multiple disciplines for well over 50 years, including in education, hiring, and machine learning.  We trace how the notion of fairness has been defined within the testing communities of education and hiring over the past half century, exploring the cultural and social context in which different fairness definitions have emerged.  In some cases, earlier definitions of fairness are similar or identical to definitions of fairness in current machine learning research, and foreshadow current formal work.  In other cases, insights into what fairness means and how to measure it have largely gone overlooked.  We compare past and current notions of fairness along several dimensions, including the fairness criteria, the focus of the criteria (e.g., a test, a model, or its use), the relationship of fairness to individuals, groups, and subgroups, and the mathematical method for measuring fairness (e.g., classification, regression). This work points the way towards future research and measurement of (un)fairness
that builds from our modern understanding of fairness while incorporating insights from the past.
\end{abstract}

\maketitle

\section{Introduction}

The United States Civil Rights Act of 1964 effectively outlawed discrimination on the basis of of an individual's race, color, religion, sex, or national origin.  The Act contained two important provisions that would fundamentally shape the public's understanding of what it meant to be {\em unfair}, with lasting impact into modern day: Title VI, which prevented government agencies that receive federal funds (including universities) from discriminating on the basis of race, color or national origin; and Title VII, which prevented employers with 15 or more employees from discriminating on the basis of race, color, religion, sex or national origin. 

Assessment tests used in public and private industry immediately came under public scrutiny. The question posed by many at the time was whether the tests used to assess ability and fit in education and employment were discriminating on bases forbidden by the new law \cite{ash1966implications}.  This stimulated a wealth of research into how to mathematically measure unfair bias and discrimination
within the educational and employment testing communities, often with a focus on race.
The period of time from 1966 to 1976 in particular gave rise to fairness research with striking parallels to ML fairness research from 2011 until today, including formal notions of fairness based on population subgroups, the realization that some fairness criteria are incompatible with one another,
and pushback on quantitative definitions of fairness due to their limitations.

Into the 1970s, there was a shift in perspective, with researchers moving from defining how a test may be {\em unfair} to how a test may be {\em fair}.  It is during this time that we see the introduction of mathematical criteria for fairness identical to the mathematical criteria of modern day.  Unfortunately, this fairness movement largely disappeared by the end of the 1970s, as the different and sometimes competing notions of fairness left little room for clarity on when one notion of fairness may be preferable to another.  Following the retrospective analysis of Nancy Cole \cite{cole2001new}, who introduced the equivalent of Hardt et al.'s 2016 equality of opportunity \cite{Hardtetal2016equality} in 1973: 
\vspace{-1em}
\begin{quote}
\fontsize{8.5}{10}\selectfont
The spurt of research on fairness issues that began in the late 1960s had results that were ultimately disappointing. No generally
agreed upon method to determine whether or not a test is fair was developed. No
statistic that could unambiguously indicate whether or not an item is fair was
identified. There were no broad technical solutions to the issues involved in
fairness. 
\end{quote}
\vspace{-.1em}
By learning from this past, we hope to avoid such a fate.

Before further diving in to the history of testing fairness, it is useful to briefly consider the structural correspondences between tests and ML models.
Test items (questions) are analogous to model features, and item responses
analogous to specific activations of those features. Scoring a test
is typically a simple linear model which produces a (possibly weighted)
sum of the item scores. Sometimes test scores are normalized or 
standardized so that scores fit a desired range or distribution.  
Because of this correspondence, much of the math is directly comparable; and many of the underlying ideas in earlier fairness work trivially map on to modern day ML fairness.  ``History doesn't repeat itself, but it often rhymes''; and by hearing this rhyme, we hope to gain insight into the future of ML fairness.

Following terminology of the social sciences, applied statistics, and
the notation of \cite{barocas2018fairness},
we use ``demographic variable'' to refer to an attribute of individuals
such as race, age or gender, denoted by the symbol $A$.
We use ``subgroup'' to denote a group of
individuals defined by a shared value of a demographic variable, e.g., $A=a$.
$Y$ indicates the ground truth or target variable, $R$ denotes a score output by a model or a test, and $D$ denotes a binary decision
made using that score. We occasionally make exceptions
when referencing original material.

\begin{figure*}[htbp]
    \centering
\begin{subfigure}{.5\textwidth}
  \centering
    \includegraphics[width=6cm]{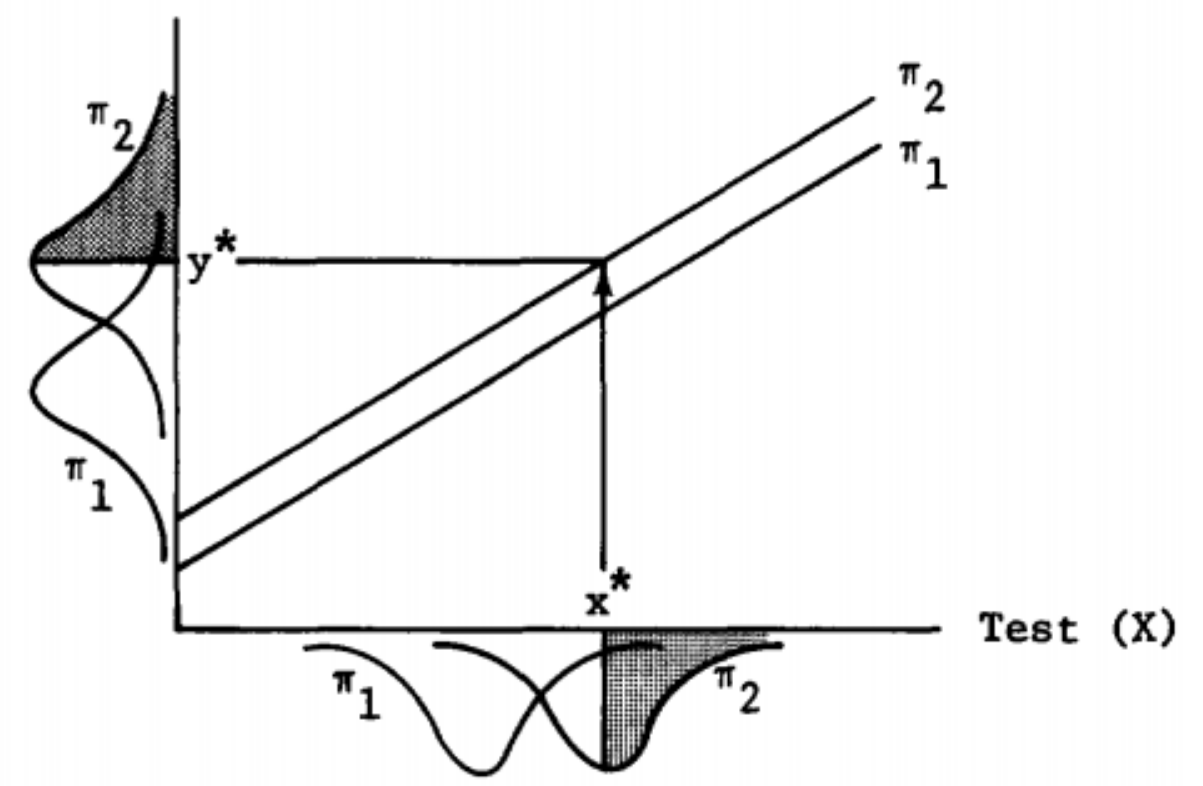}
    \caption{Labels on regression lines indicate which subgroup they fit.}
  \label{fig:cleary_a}
\end{subfigure}%
\begin{subfigure}{.5\textwidth}
  \centering
    \includegraphics[width=6cm]{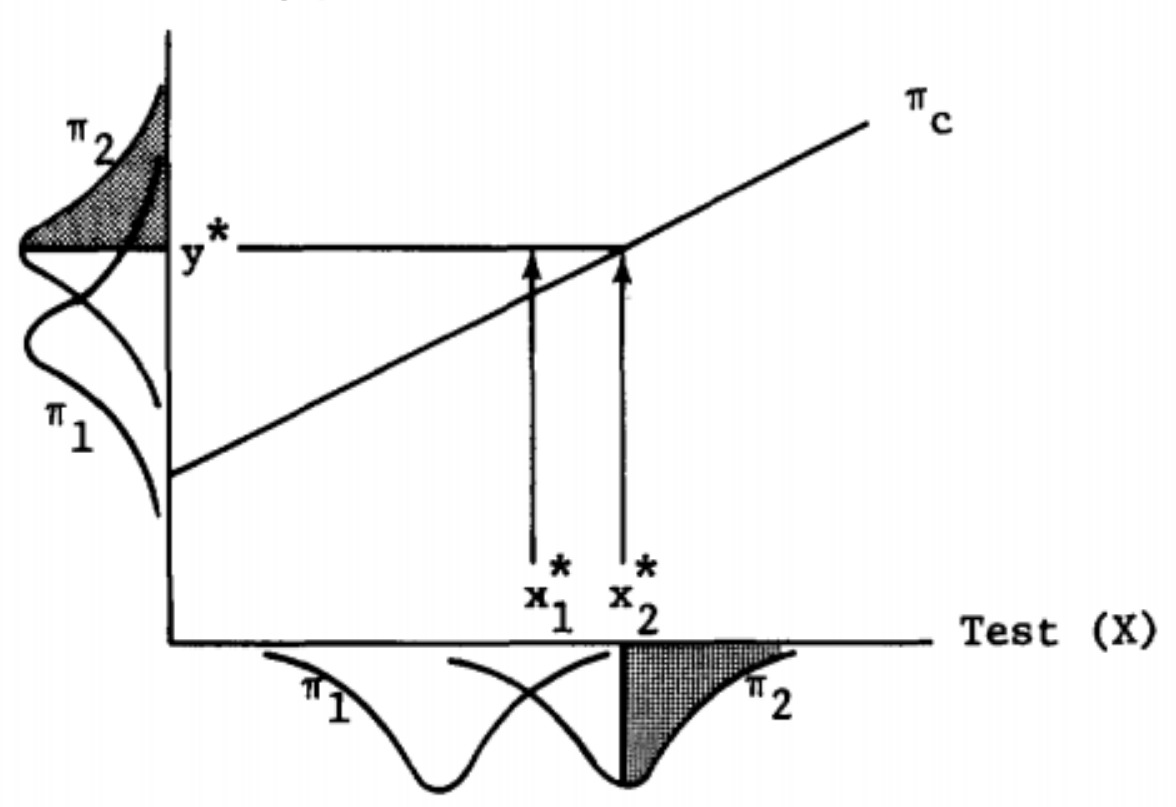}
    \caption{The regression line labeled $\pi_c$ fits both subgroups
    separately (and hence also their union).}\vspace{-1em}
  \label{fig:cleary_b}
\end{subfigure}%
\caption{Petersen and Novick's \cite{petersen1976evaluation} original
figures demonstrating fairness criteria.
The marginal distributions of test scores and ground truth
scores for subgroups $\pi_1$ and $\pi_2$ are shown by the axes.
}\label{fig:cleary}
\end{figure*}

\section{History of fairness in testing}\label{sec:history}

\subsection{1960s: Bias and Unfair Discrimination}\label{sec:60s}

Concerned with the fairness of tests for black and white students,
T.\ Anne Cleary defined a quantitative measure of test bias for the first time, cast in terms of a formal model for predicting educational
outcomes from test scores \cite{cleary1966test, cleary1968test}:\vspace{-.15em}
\begin{quote}
\fontsize{8.5}{10}\selectfont
A test is {\em biased} for members of a subgroup of the population if, in the prediction of a criterion
for which the test was designed, consistent nonzero errors of prediction are made for members of the subgroup. In other words, the test is biased if the criterion score predicted from
the common regression line is consistently too high or too low for members of the subgroup.
With this definition of bias, there may be a connotation of ``unfair," particularly if the use of
the test produces a prediction that is too low. (Emphasis added.)
\end{quote} 

According to Cleary's criterion, the situation depicted in Figure~\ref{fig:cleary_a} is biased for members of subgroup $\pi_2$ if the regression line $\pi_1$
is used to predict their ability, since it underpredicts
their true ability. For Cleary, the situation depicted in Figure~\ref{fig:cleary_b} is not biased: since data from each of the subgroups produce the same regression line,
that line can be used to make predictions for either group.

In addition to defining bias in terms of predictions
by regression models, Cleary also performed a study on real-world data from three state-supported and state-subsidized schools, comparing college GPA with SAT
scores.
Racial data was obtained from an admissions office, from an NAACP list of black students, and from examining class pictures.
Cleary used Analysis of Covariance ({\sc ANCOVA}) to test the relationships between
SAT and HSR scores with GPA grades.
Contrary to some expectations, Cleary found little evidence of the SAT being
a biased predictor of GPA. (Later, larger studies found that the SAT
overpredicted the GPA of black students \cite{vars1998scholastic}; 
it may be that the SAT is biased but less so than the GPA.)

While Cleary's focus was on education, her contemporary Robert Guion was 
concerned with unfair discrimination in employment. 
Arguing for the importance of quantitative analyses in 1966, he wrote that: ``Illegal discrimination is largely an ethical matter, but the fulfillment of ethical responsibility begins with technical competence'' \cite{guion1966employment}, 
and defined 
unfair discrimination to be 
``when persons with equal probabilities of success on the job have unequal probabilities of being hired for the job.'' However, Guion recognized the challenges in using constructs such as the probability of success. We can observe actual success and failure
after selection, but the probability of success is not itself observable, and  a sophisticated model is required to estimate it at the time of selection. 

By the end of the 1960s, there was political and legal support
backing concerns with the unfairness of the educational system for black children and the unfairness of tests
purporting to measure black intellectual competence.  Responding to these concerns, the Association of Black Psychologists formed in 1969 
immediately published ``A Petition of Concerns'',
 calling for a moratorium on standardized tests ``(which are used) to 
maintain and justify the practice of systematically denying
economic opportunities''
\cite{williams1980war}. 
The NAACP followed up on this in 1974 by adopting a resolution that demanded ``a moratorium on standardized testing wherever such tests have not been corrected for cultural bias'' (cited by \cite{samuda1998psychological}).
Meanwhile, advocates of testing
worried that alternatives to testing such as interviews
would introduce more subjective bias \cite{flaugher1974bias}.\footnote{
For example, the origins of the college entrance essay are rooted
in ivy league universities' covert attempts to suppress the numbers of Jewish students, whose performance on
entrance exams had led them to become an increasing
percentage of the student population \cite{karabel2006chosen}.}

\subsection{1970s: Fairness}\label{sec:70s}

As the 1960s turned to the 1970s, work began to arise that parallels the recent evolution of work in ML fairness, marking a change in framing from {\em unfairness} to {\em fairness}. 
Following Thorndike \cite{thorndike1971concepts}, ``The discussion of `fairness' in what has gone before is clearly over-simplified.
In particular, it has been based upon the premise that the available criterion score is a perfectly relevant, reliable and unbiased measure...''  Thorndike's  sentiment was shared by other academics of the time, who, in examining the earlier work of Cleary, objected that it failed to take into account the
differing false positive and false negative rates that occur
when subgroups have different base rates (i.e., $A$ is not independent of $Y$) \cite{thorndike1971concepts,einhorn1971methodological}.

With the goal of moving beyond simplified models, 
Thorndike \cite{thorndike1971concepts} proposed one of the first quantitative criteria for measuring test {\em fairness}.  With this shift, Thorndike advocated for considering the {\em contextual use} of a test: \vspace{-.15em}
\begin{quote}
\fontsize{8.5}{10}\selectfont
A judgment on test-fairness must rest on the inferences that are made from the test rather than on a comparison of mean scores in the two populations. One must then focus attention on fair use of the test scores, rather
than on the scores themselves.
\end{quote}

Contrary to Cleary, Thorndike argued that sharing a common regression line
is not important, as one can achieve fair selection goals by using different
regression lines and different selection thresholds for the two groups.

As an alternative to Cleary, Thorndike proposed that the ratio of predicted positives to ground truth positives be equal for each group. Using confusion matrix terminology, this is equivalent to requiring that the ratio
$(TP+FP)/(TP+FN)$ be equal for each subgroup. 
According to Thorndike, the situation in Figure~\ref{fig:cleary_a} is fair for 
test cutoff $x^*$. Figure~\ref{fig:cleary_b} is unfair using any single threshold, but fair if threshold $x_1^*$ is used for group $\pi_1$ and threshold
$x_2^*$ is used for group $\pi_2$. 

Similar to modern day ML fairness, e.g., Friedler et al.\ in 2016 \cite{friedler2016possibility}, Thorndike also pointed out the tension between individual notions of fairness and group notions of fairness: ``the two definitions of fairness---one based on predicted criterion score for individuals and the other on the distribution of criterion scores in the two groups---will always be in conflict.'' The conflict was also raised by others in the period, including 
Sawyer et al.\ \cite{sawyer1976utilities}, in a foreshadowing
of the {\sc compas} debate of 2016: 
\begin{quote}
\fontsize{8.5}{10}\selectfont
A conflict arises because the success maximization procedures based on individual
parity do not produce equal opportunity (equal selection for equal success) based on
group parity and the opportunity procedures do not produce success maximization
(equal treatment for equal prediction) based on individual parity.
\end{quote}

Almost as an aside, Thorndike mentions the existence of another
regression line ignored by Cleary: the line that estimates the value of 
the test score
$R$ given the target variable $Y$. This idea hints at the notion of equal opportunity
for those with a given value of $Y$, an idea which soon was picked up by Darlington \cite{darlington1971another} and Cole \cite{cole1973bias}.

At a glance, Cleary's and Thorndike's definitions are difficult to
compare directly because of the different ways in which they're defined.
Darlington \cite{darlington1971another} helped to shed light on the relationship
between Cleary and Thorndike's conceptions of fairness by expressing them
in a common formalism. He defines four fairness criteria in terms
of the correlation $\rho_{AR}$ between the demographic variable and the
test score.  Following Darlington,

\begin{enumerate}
\item Cleary's criterion can be restated in terms of 
correlations of the ``culture variable'' with test scores.  If Cleary's criterion holds for every subgroup, then  $\rho_{AR} = \rho_{AY}/\rho_{RY}$
 \cite{vargha1996dichotomization}.
 \footnote{\label{foot:normal}Although Darlington does not mention this additional constraint, we believe the criterion only holds if $A$, $R$ and $Y$ have a 
multivariate normal distribution.}

\item Similarly, Thorndike's criterion is 
equivalent to requiring that $\rho_{AR} = \rho_{AY}$.
\item The
criterion $\rho_{AR} = \rho_{AY}\times\rho_{RY}$ is 
motivated by thinking about $R$ as a dependent variable 
affected by independent variables $A$ and $Y$. 
If $A$ has no direct effect on $R$ once $Y$ is taken
into account then we have a zero partial correlation,
i.e.\ $\rho_{AR.Y}=0$.\footnote{See footnote \ref{foot:normal}
}.
\item An alternative ``starkly simple'' criterion of $\rho_{AR} = 0$ (recognizable as modern day demographic parity \cite{DworkEtAl2012}) is introduced but not dwelt on.
\end{enumerate}

\begin{figure}
    \centering
    \includegraphics[width=5cm]{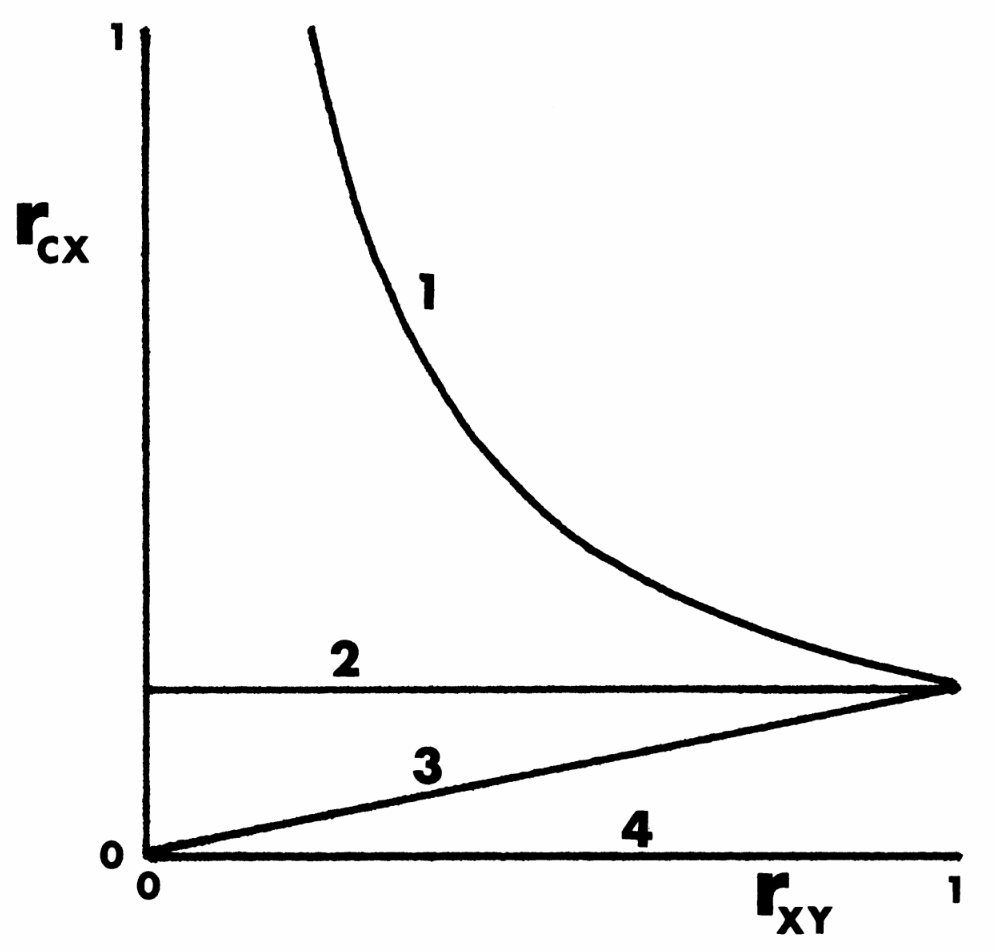}
    \vspace{-.5em}
    \caption{Darlington's original graph of fair values of the correlation between culture and test score ($r_{CX}$ in Darlington's notation), plotted against the correlation between test score and ground truth
    ($r_{XY}$), according to his definitions (1--4). (The correlation between the
    demographic and target variables is assumed here to be fixed at 0.2.)}
    \label{fig:darlington_graph}
    \vspace{-1em}
\end{figure}

Darlington's mapping of Cleary's and Thorndike's criteria lets him
prove that they're incompatible except in the special cases where
the test perfectly predicts the target variable ($\rho_{RY}=1$), or where
the target variable is uncorrelated with the demographic variable
($\rho_{AY}=0$). Figure~\ref{fig:darlington_graph},
 reproduced from Darlington's 1971 work, shows that, for
any given non-zero correlation between the demographic and  
target variables, definitions (1), (2), and (3) converge as the correlation
between the test score and the target variable approach 1. When the
test has only a poor correlation with the target variable, there may be
no fair solution using definition (1).

Figure~\ref{fig:darlington_graph} enables a
range of further observations. According to definition (1), 
for a given correlation between demographic and
target variables, the lower the correlation of the test with the target
variable, the higher it is allowed to correlate with the demographic
variable and still be considered fair. 
Definition (3), on the other hand, is the opposite, in that 
the lower the correlation of the test with the target
variable, the lower too must be the the test's correlation with the demographic
variable.
Darlington's criterion (2) is the geometric mean of criteria (1) and (3):
``a compromise position midway between [the] two... however, a compromise may
end up satisfying nobody; psychometricians are not in the habit of agreeing
on important definitions or theorems by compromise.'' Darlington 
shows that definition (3) is the only one of the four whose
errors are uncorrelated with the demographic variable, where by ``errors'', 
he means errors in the regression task of estimating $Y$ from $R$.

In 1973, Cole \cite{cole1973bias} continued exploring ideas of equal outcomes across subgroups, defining fairness as all subgroups having the same True Positive 
Rate (TPR), recognizable as modern day equality of opportunity \cite{Hardtetal2016equality}.
That same year, Linn \cite{linn1973fair} introduced (but did not advocate for) 
equal Positive Predictive Value (PPV) as a fairness criterion, recognizable as modern day predictive parity \cite{chouldechova2017fair}.\footnote{Although he cites \cite{guion1966employment}
and \cite{einhorn1971methodological}, a seeming misattribution, as pointed out by
\cite{petersen1976evaluation}.}

Under Cleary and Darlington's conceptions,  
bias or (un)fairness is a property of the test itself.  This is contrary to Thorndike, Linn and Cole, who take fairness to be a property of the use of a test. 
The latter group tended to assume that a test is static, and focused on optimizing
 its use; whereas Cleary's concerns were with how to improve the tests
themselves. Cleary worked for Educational Testing Services, and one
can imagine a test being designed allowing for a range of use cases,
since it may not be knowable in advance either
i) the precise populations on which it will be deployed, nor
ii) the number of students which an institution deploying the test
is able to offer places to. 

By March  1976, the interest in fairness in the educational testing community was
so strong that an entire issue of the Journal of Education Measurement was devoted
to the topic \cite{jem_13_1}, including a lengthy lead article 
by Peterson and Novick \cite{petersen1976evaluation}, in which they consider for the first time the equality of
True Negative Rates (TNR) across subgroups, and  equal TPR / equal TNR across subgroups (modern day equalized odds \cite{Hardtetal2016equality}).
Similarly, they consider the case of equal PPV and equal NPV across subgroups.\footnote{
They do not advocate for either combination (neither equal TPR and TNR, nor 
equal PPV and NPV) on the grounds that
either combination requires unusual circumstances. However there is a flaw in their reasoning. For example, arguing against equal TPR and equal TNR, they claim that this requires equal base rates in the ground truth in addition to equal TPR.}

Work from the mid-1960s to mid-1970s
can be summarized along four distinct categories: {\sc individual}, {\sc non-comparative}, {\sc subgroup parity}, and {\sc correlation}, defined in Table~\ref{tab:criteria_categories}.
 It should be emphasized that in not all cases where a researcher defined a
criterion did they also advocate for it. In particular, Darlington, Linn, Jones, 
and Peterson and Novick all define criteria purely for the purposes of exploring
the space of concepts related to fairness.  A summary of fairness technical definitions during this time is listed in Table \ref{tab:fairness_criteria_summary}.

\begin{table}
\begin{tabular}{lp{5cm}}
\toprule
\bf Category & \bf Description \\
\bottomrule
{\sc individual} & Fairness criterion defined purely in terms of individuals \\
\hline
{\sc non-comparative} & Fairness criterion for each subgroup does not reference other subgroups \\
\hline
{\sc subgroup parity} & Fairness criterion defined in terms of parity
of some value across subgroups \\
\hline
{\sc correlation} & Fairness criterion defined in terms of the
correlation of the demographic variable with the model output \\
\bottomrule
\end{tabular}
\caption{Categories of Fairness Criteria}
\label{tab:criteria_categories}
\vspace{-2em}
\end{table}

\subsection{Mid-1970s: The Fairness Tide Turns}

Immediately after the the journal issue of 1976, research into quantitative definitions of test fairness
seems to have come to a halt.  Considering why this happened may be a valuable lesson to learn from for modern day fairness research. The same Cole who in 1973 proposed equality of TPR, wrote in 2001 that \cite{cole2001new}:

\begin{quote}
\fontsize{8.5}{10}\selectfont
In short, research over the last 30 or so years has not supplied any analyses
to unequivocally indicate fairness or unfairness, nor has it produced clear procedures to avoid unfairness. To make matters worse, the views of fairness of the
measurement profession and the views of the general public are often at odds.
\end{quote}

Foreshadowing this outcome, statements from researchers in the 1970s indicate an  increasing concern with how fairness criteria obscure ``the fundamental problem, which is to find some rational basis for providing compensatory
treatment for the disadvantaged'' \cite{novick1976towards}.  Following Peterson and Novick, the concepts of culture-fairness and group parity are not viable in practice, leading to models that can sanction the discrimination they seek to rectify \cite{petersen1976evaluation}.  They argue that fairness should be reconceptualized as a problem in maximizing expected utility \cite{petersen1976expected}, recognizing  ``high social utility in equalizing opportunity and 
reducing disadvantage'' \cite{novick1976towards}.

A related thread of work highlights that different fairness criteria encode 
different value systems \cite{hunter1976critical}, and that 
quantitative techniques alone cannot answer
the question of which to use.
In 1971, Darlington \cite{darlington1971another} urges that the concept of ``cultural fairness'' be replaced by ``cultural
optimality'', which takes into account a policy-level question concerning the optimum balance between accuracy and cultural factors.  In 1974, Thorndike points out that ``one's value system is deeply involved in one's judgment as to what is `fair use'
of a selection device'' \cite{novick1976towards}),  
and similarly, in 1976, Linn \cite{linn1976insearch} draws attention to the fact that ``Values are implicit in the models. To adequately address issues of values they need to be dealt with explicitly.'' Hunter and Schmidt \cite{hunter1976critical} begin to address this issue by bringing ethical theory to the discussion, relating fairness to theories of individualism and proportional representation.  Current work may learn from this point in history by explicitly connecting fairness criteria to different cultural and social values.

\begin{table*}
\renewcommand{\arraystretch}{1.25}
\raggedleft
\begin{adjustwidth}{-.075em}{}
\begin{tabular}{@{\hspace{.5em}}p{3.3cm}p{6.6cm}lp{4.2cm}}
\toprule
\bf Source &\bf Criterion & \bf Category  & \bf Proposition \\
\bottomrule
\multirow{2}{*}{Guion (1966)} &
``people with equal probabilities of success on the job have equal probabilities of being hired for the job'' & \multirow{2}{*}{{\sc individual}} &  \multirow{2}{11em}{Is the {\bf use} of the {\bf test} fair?}\\
\hline
Cleary (1966) & ``a subgroup does not have consistent errors'' & {\sc non-comparative} & \multirow{1}{*}{Is the {\bf test} fair to {\bf subgroup} $a$?} \\\hline
\multirow{2}{*}{Einhorn and Bass (1971)$^\dagger$}  &
$Prob(Y>y* | R=r_a*, A=a)$ is constant for all subgroups $a$
& \multirow{2}{*}{{\sc subgroup parity}} & \multirow{2}{12em}{Is the {\bf use} of the {\bf test} fair with respect to $A$?} \\\hline
\multirow{2}{*}{Thorndike (1971)} &
$Prob(R>=r_a* | A=a) / Prob(Y>=y* | A=a)$ is constant for all subgroups $a$ &
\multirow{2}{*}{{\sc subgroup parity}} & \multirow{2}{12em}{Is the {\bf use} of the {\bf test} fair with respect to $A$?} \\\hline
Darlington (1971) (1) &
$\rho_{AX} = \rho_{AY}/\rho_{RY}$ (equivalent to $\rho_{AY.R} = 0$) &
\multirow{4}{6em}{{\sc correlation}} & \multirow{4}{14em}{Is the {\bf test}
fair with respect to $A$?} \\
Darlington (1971) (2) &
$\rho_{AR} = \rho_{AY}$ &
  &  \\
Darlington (1971) (3) &
$\rho_{AR} = \rho_{AY}\times\rho_{RY}$ (equivalent to $\rho_{AR.Y} = 0$) &
  & \\
Darlington (1971) (4) &
$\rho_{AR} = 0$ &  & \\
Darlington (1971)  &
\multirow{2}{22em}{$\rho_{R(Y-kA)}$, is maximized where $k$ is the subjective value placed on subgroup attribute $A=1$} &
\multirow{2}{*}{{\sc correlation}} & Does the {\bf test} produce the   \\
{\em culturally optimum$^\dagger$} & & & optimal outcome w.r.t.\ $A$?\\
\hline
\multirow{2}{*}{Cole (1973)} &
$Prob(R>=r_a* | Y>=y*, A=a)$ is constant for all subgroups $a$ &
\multirow{2}{*}{{\sc subgroup parity}} & \multirow{2}{12em}{Is the {\bf use} of the {\bf test} fair with respect to $A$?} \\
\hline
\multirow{2}{*}{Linn (1973)} &
$Prob(Y>=y* | R>=r_a*, A=a)$ is constant for all subgroups $a$ &
\multirow{2}{*}{{\sc subgroup parity}} & \multirow{2}{12em}{Is the {\bf use} of the {\bf test} fair with respect to $A$?} \\
\hline
Jones (1973) & \multirow{2}{*}{$E(\hat{Y}|a) = E(Y|a)$} & \multirow{2}{*}{{\sc non-comparative}} & \multirow{2}{*}{Is the {\bf test} fair to {\bf subgroup} $a$?} \\ 
{\em mean fair}$^\dagger$ & & & \\

Jones (1973) & \multirow{3}{20em}{a subgroup $a$ has equal representation in the top-$n$ candidates ranked by model score as it has in the top-$n$ candidates ranked by $Y$, for all $n$} & \multirow{3}{*}{{\sc non-comparative}} & \multirow{3}{*}{Is the {\bf test} fair to {\bf subgroup} $a$?} \\
{\em general standard}$^\dagger$ & & & \\ & & & \\
Jones (1973)  &
\multirow{3}{20em}{a subgroup $a$ has equal representation in the top-$n$ candidates ranked by model score as it has in the top-$n$ candidates ranked by $Y$} &
{\sc non-comparative} & \multirow{3}{12em}{Is the {\bf use} of the {\bf test} fair to {\bf subgroup} $a$?} \\
{\em at position $n$}$^\dagger$ & & & \\ & & & \\
\hline

Peterson~\&~Novick~(1976)  &
\multirow{3}{22em}{$Prob(R>=r_a* | Y>=y*, A=a)$ is constant for all subgroups $a$, and $Prob(R<r_a* | Y<y*, A=a)$ is constant for all subgroups $a$} &
\multirow{3}{*}{{\sc subgroup parity}} & \multirow{3}{12em}{Is the {\bf use} of the {\bf test} fair with respect to $A$?} \\
{\em conditional probability and its converse} & & & \\
Peterson~\&~Novick~(1976)  &
\multirow{3}{22em}{$Prob(Y>=y* | R>=r_a*, A=a)$ is constant for all subgroups $a$, and $Prob(Y<y* | R<r_a*, A=a)$ is constant for all subgroups $a$} &
\multirow{3}{*}{{\sc subgroup parity}} & \multirow{3}{12em}{Is the {\bf use} of the {\bf test} fair with respect to $A$?} \\
{\em equal probability and its converse} &&& \\
\bottomrule
\end{tabular}
\end{adjustwidth}

\caption{Early technical definitions of fairness in educational and employment testing. Variables: $R$ is the test score; $Y$ is
the target variable; $A$ is the demographic variable. The Proposition column indicates whether fairness is considered a property of the way in
which a test is used, or of the test itself.
$^\dagger$ indicates that the criterion is discussed in the appendix.} 
\label{tab:fairness_criteria_summary}
\vspace{-2em}
\end{table*}

\subsection{1970s on:  Differential Item Functioning}\label{sec:60s_70s_diff}

Concurrent with the development of criteria for the fair use of tests, another line of research in the measurement community concerned 
looking for bias in test questions (``items'').
In 1968, Cleary and Hilton \cite{cleary1968investigation} 
used an analysis of variance ({\sc ANOVA}) design to 
test the interaction between race, socioeconomic level
and test item.  Ten years later, the related idea of Differential Item Functioning (DIF) was introduced by Scheuneman in 1979 \cite{scheuneman1979method}:
``an item is considered unbiased if, for persons with
the same ability in the area being measured, the probability of a correct
response on the item is the same regardless of the population group membership of the individual.'' 
That is, if $I=I(q)$ is the variable representing a 
correct response on question $q$, then by this definition $I$
is unbiased if  $A \perp I | Y$.

\begin{figure}
    \centering
    \includegraphics[width=4cm]{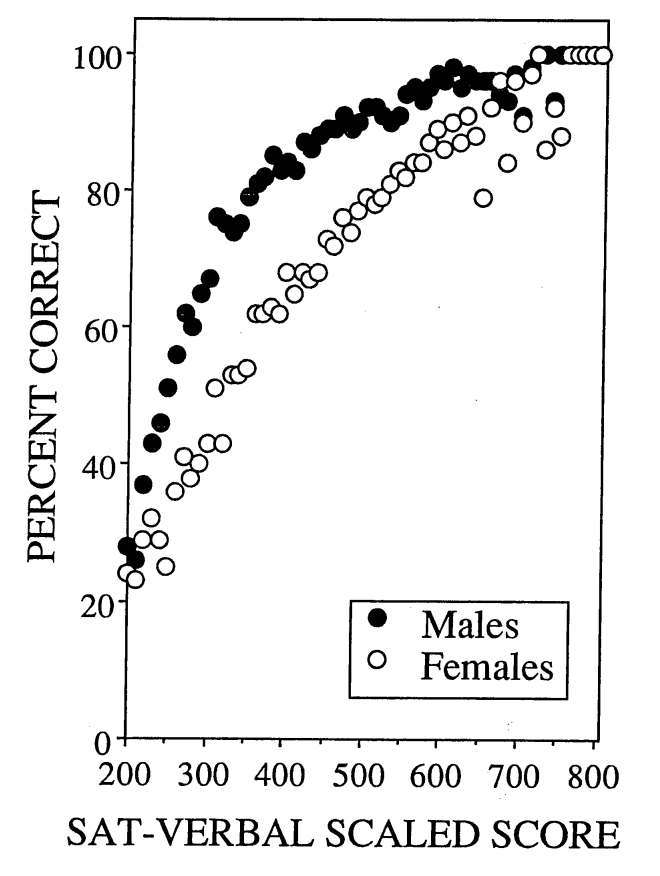}
        \vspace{-1em}
    \caption{Original graph from \cite{dorans1992dif} illustrating DIF.}
    \vspace{-.5em}
    \label{fig:DIF}
\end{figure}

In practice, the best measure of the ability that the item is testing
is often the test in which the item is a component \cite{dorans2017contributions}:
\begin{quote}
\fontsize{8.5}{10}\selectfont
A major change from focusing primarily on fairness in a domain, where so many factors could spoil the validity effort, to a domain where analyses could be conducted in a relatively simple, less confounded way. ... In a DIF analysis, the item is evaluated against something designed to measure a particular construct and something that the test producer controls, namely a test score.
\end{quote}
Figure~\ref{fig:DIF} illustrates DIF for a test item.

DIF became very influential in the education field, and
to this day DIF is in the toolbox of test designers. 
Items displaying a DIF are ideally examined further to identify
the cause of bias, and possibly removed from the test \cite{penfield2016fairness}.

\subsection{1980s and beyond}

With the start of the 1980s came renewed public debate 
about the existence of racial differences in general intelligence,
and the implications for fair testing,
following the publication of the controversial {\em Bias in Mental Testing}
\cite{jensen1980bias}. Political opponents of group-based 
considerations in educational and employment practices 
framed them in terms of ``preferential treatment''
for minorities and ``reverse discrimination'' against whites.
Despite, or perhaps because of, much public debate,
neither Congress nor the courts gave unambiguous answers to the question
of how to balance social justice considerations with the historical
and legal importance placed on the individual in the United States \cite{national1989fairness}.

Into the 1980s, courts were asked to rule on many cases involving (un)fairness in 
educational testing. 
To give just one example, 
Zwick and Dorans \cite{zwick2016philosophical} described the case of Debra P.\ v.\ Turlington  1984, in which a lawsuit was filed on behalf of ``present and future twelfth grade students who had failed or would fail'' a high school graduation test. The initial ruling found that the test perpetuated past discrimination and was in violation of the Civil Rights Act.  More examples of court rulings on fairness are given by  \cite{phillips2016legal, zwick2016philosophical}.

By the early 1980s, ideas about fairness were having a widespread
influence on U.S.\ employment practices. In 1981, with no
public debate, the United States
Employment Services implemented score-adjustment strategy that was
sometimes called ``race-norming'' \cite{rice1994race}. Each individual is assigned
a percentile ranking within their own ethnic group, rather than
to the test-taking population. By the mid-1980s, race-norming was ``a highly
controversial issue sparking heated debate.'' The debate was settled through
legislation, with the 1991 Civil Rights Act banning the practice of race-norming
\cite{west2011fairness}.

\section{Connections to ML fairness}

\subsection{Equivalent Notions}\label{sec:equivalent_notions}

Many of the fairness criteria we have overviewed are identical to modern-day fairness definitions.  Here is a brief summary of these connections:

\begin{itemize}
    \item Peterson and Novick's ``conditional probability and its converse''
is equivalent to what in ML fairness is variously called {\em sufficiency} \cite{barocas2018fairness},
{\em equalized odds} \cite{Hardtetal2016equality}, or {\em conditional procedure accuracy} \cite{berk2017fairness}, sometimes expressed as the conditional independence 
$A \perp D | Y$. 
\item Similarly, their
``equal probability and its converse'' is equivalent to what is called
{\em sufficiency} \cite{barocas2018fairness} or {\em conditional use accuracy equality} \cite{berk2017fairness}, $A \perp Y | D$.
\item Cole's 1973 fairness definition is identical to {\em equality of opportunity}
\cite{Hardtetal2016equality}, $A \perp D | Y=1$.
\item Linn's 1973 definition is equivalent
to {\em predictive parity}  \cite{chouldechova2017fair}, $A \perp Y |D=1$.
\item Darlington's criterion (1) is equivalent to
sufficiency in the special case where $A$, $R$ and $Y$
have a multivariate Gaussian distribution.
This is because
for this special case the partial correlation
$\rho_{AY.X} = 0$ is equivalent to
$A\perp Y |R$ \cite{baba2004partial}. In general though, 
we cannot assume even a one way implication, since 
$A\perp Y |R$ does not imply $\rho_{AY.X} = 0$ (see \cite{vargha1996dichotomization} for a counterexample).
\item Similarly, Darlington's criteria (2) and (3) are equivalent to
independence and separation only in the special cases of
multivariate Gaussian  distributions.
\item Darlington's definition (4) is a relaxation of what is called
{\em independence} \cite{barocas2018fairness} or {\em demographic parity} in 
ML fairness, i.e. $A \perp R$;
it is equivalent when $A$ and $R$ have a bivariate Gaussian distribution.

\item Guion's definition ``people with equal probabilities of success on the job
have equal probabilities of being hired for the job'' is
a special case of Dwork's \cite{DworkEtAl2012} {\em individual fairness} with the presupposition that ``probability of success on the job'' is a construct that
can be meaningfully reasoned about. 
\end{itemize}

The fairness literature in both the fields of ML and in testing have also been motivated by causal
considerations \cite{kusner2017counterfactual, Hardtetal2016equality}. Darlington 
\cite{darlington1971another} motivate his definition (3) on the basis
of a causal relationship between $Y$ and $R$ (since
an ability being measured affects the performance on the test).
However \cite{hunter1976critical}
have pointed out that in testing scenarios we typically only have a proxy
for ability, such as later GPA 4 years later, and it is wrong to draw
a causal connection from GPA to college entrance exam.

Hardt et al.\ \cite{Hardtetal2016equality} describe the challenge in 
building causal models, by considering two distinct models and their consequences and concluding 
that ``no test based only on the target labels, the protected attribute and the score would give
different indications for the optimal score $R^{*}$ in the two scenarios.''
This is remarkably reminiscent of Anastasi \cite{anastasi1961psychological}, writing in 1961 about test fairness:
\begin{quote}
\fontsize{8.5}{10}\selectfont
No test can eliminate
causality. Nor can a test score, however derived, reveal the
origin of the behavior it reflects. If certain environmental
factors influence behavior, they will also influence those
samples of behavior covered by tests. When we use tests to
compare different groups, the only question the tests can
answer directly is: ``How do these groups differ under existing
cultural conditions?''
\end{quote}

Both the testing fairness and ML fairness literatures have also paid
great attention to impossibility results, such as the 
distinction between group fairness and individual fairness, and
the impossibility of obtaining more than one of separation,
sufficiency and independence except under special conditions
\cite{thorndike1971concepts,darlington1971another,petersen1976evaluation,
barocas2018fairness, chouldechova2017fair,kleinberg2016inherent}.

In addition, we see some striking parallels in the framing of fairness in terms 
of ethical theories, including explicit advocacy for utilitarian approaches.  
\begin{itemize}
\item Petersen and Novick's utility-based approaches relate to Corbett-Davies et al.'s 
framing of the cost of fairness \cite{CorbettDavies2017EtAl}.
\item Hunter and Schmidt's analysis of the value systems underlying fairness criteria 
is similar in spirit to Friedler et al.'s relation of fairness criteria and different worldviews
\cite{friedler2016possibility}.
\end{itemize}

\subsection{Variable Independence}

As briefly mentioned above, modern day ML fairness has categorized fairness definitions in terms of independence of variables, which includes {\em sufficiency} and {\em separation} \cite{barocas2018fairness}. Some historical notions of fairness neatly fit into this categorization, but others shed light on further dimensions of fairness criteria.  Table~\ref{tab:fairness_relationships} summarizes these connections, linking the historical criteria
introduced in Section~\ref{sec:history} to modern day categories. (Utility-based criteria are omitted, but will be discussed below.)  

\begin{table*}
\renewcommand{\arraystretch}{1.15}
\small
\begin{tabular}{p{5cm}p{3.25cm}p{8.5cm}}
\toprule
\bf Historical criterion & \bf ML fairness criterion & \bf Relationship \\
\bottomrule
Guion (1966) & {\sc individual} & relaxation \\
\hline
\multirow{2}{*}{Cleary (1968)} & \multirow{2}{*}{{\sc sufficiency}} & when Cleary's criterion holds for
all subgroups then we we have equivalence when $R$ and $Y$ have
bivariate Gaussian distribution\\
\hline
\multirow{2}{*}{Einhorn and Bass (1971)} &
\multirow{2}{*}{{\sc sufficiency}} &
both involve probability of $Y$ conditioned on $R$, but 
Einhorn and Bass are only concerned with the conditional likelihood
at the decision threshold\\
\hline
Thorndike (1971) & --- & --- \\
\hline
Darlington (1971) (1) & {\sc sufficiency} &
equivalent when variables have a multivariate Gaussian  distribution \\
Darlington (1971) (2) & --- & --- \\
Darlington (1971) (3) & {\sc separation} & equivalent when variables have a multivariate Gaussian  distribution\\
Darlington (1971) (4) & {\sc independence} & equivalent when variables have a bivariate Gaussian  distribution\\
\hline 
Cole (1973) &
{\sc separation} & relaxation (equivalent to equality of opportunity) \\
\hline
Linn (1973) & {\sc sufficiency} & relaxation (equivalent to predictive parity)\\
\hline
Jones (1973) {\em mean fair} & --- & --- \\
Jones (1973) {\em at position $n$} & --- & --- \\
Jones (1973) {\em general criterion} & --- & --- \\
\hline
Peterson and Novick (1976)  & \multirow{2}{*}{{\sc separation}} & \multirow{2}{*}{equivalent} \\
{\em conditional probability and its converse} && \\
Peterson and Novick (1976)  & \multirow{2}{*}{{\sc sufficiency}} & \multirow{2}{*}{equivalent} \\
{\em equal probability and its converse} && \\
\bottomrule
\end{tabular}
\caption{Relationships between testing criteria and ML's independence criteria}
\label{tab:fairness_relationships}
\vspace{-1.25em}
\end{table*}

We find that non-comparative criteria (discussed by Cleary and Jones) do not map
onto any of the independence
conditions used in ML fairness. Similarly, Thorndike's,
and Darlington's  have no counterparts that we know of.
There are conceptual similarities between Jones' criteria 
and the constrained ranking problem described by \cite{celis2017ranking},
and also between Einhorn's criterion and  concerns 
about infra-marginality \cite{simoiu2017problem}.

For a binary classifier,
Thorndike's 1971 group parity criterion is equivalent to requiring that the ratio
of positive predictions to ground truth positives be equal for all subgroups.
This ratio has no common name that we could find
(unlike e.g., precision, recall, etc.), although \cite{petersen1976evaluation} refer to this as the ``Constant Ratio Model''.  
It is closely related to coverage constraints \cite{goh2016satisfying},
class mass normalization
\cite{zhu2003semi} and expectation regularization \cite{mann2007simple}.
Similar arguments can be made for Darlington's criterion (2) and Jones' criteria ``at position $n$'' and ``general
criterion''. When viewed as a model of subgroup quotas
\cite{hunter1976critical}, Thorndike's criterion is reminiscent of {\it fair division} in economics.

\subsection{Regression and Correlation}

In reviewing the history of fairness in testing, it becomes clear that regression models have played a much larger role than in the ML community. Similarly, the use of correlation as a fairness criterion is all but absent in modern ML
Fairness literature.

Given that correlation of two variables is a weaker criterion than independence,
it is reasonable to ask why one might want a fairness criterion defined in terms
of correlations. One practical reason is that calculating correlations is a lot
easier than estimating independence. Whereas correlation is a descriptive statistic,
and so calculating requires few assumptions, estimating independence requires an
the use of inferential statistics, which can in general be highly non-trivial
\cite{shah2018hardness}.

Considering the analogy between model features and test items described
in the Introduction, we also know of no ML analogs to the Differential Item Functioning. Such analogs
might test for bias in model features. Instead, one approach adopted in 
ML fairness has been the use of adversarial methods to mitigate the effects
of features with undesirable correlations with subgroups, e.g., \cite{beutel2017,ZhangEtAl2018}.

\subsection{Model vs.~Model Use}

Section~\ref{sec:history} described how the test literature had competing
notions of whether fairness is a property of a {\em test}, or of the
{\em use of a test}. 
A similar discussion of whether ML models can be judged as fair
or unfair independent of a specific use (including a specific
model threshold) has been largely implicit or missing in
the ML fairness literature. Models are sometimes trained
to be ``fair'' at their default decision threshold (e.g., 0.5), although
the use of different thresholds can have a major impact on fairness
\cite{Hardtetal2016equality}. The ML fairness notion of {\em calibration}, i.e., $P(Y=1 | A=a, R=r)=r$ for all $a$ and $r$,
can be interpreted to be a property of the model rather than of its use,
since it does not depend on 
the choice of decision threshold. 

\subsection{Race and Gender}
Some work on practically assessing fairness in ML has tackled the problem of using
race as a construct. This echoes concerns in the testing literature that stem back
to at least 1966: ``one stumbles immediately over the scientific difficulty of establishing clear yardsticks by which people can be classified into convenient racial categories''
\cite{guion1966employment}. Recent approaches have used Fitzpatrick skin type or unsupervised clustering to avoid racial categorizations \cite{buolamwini2018gender, ryuinclusivefacenet}.
We note that the testing literature of the 1960s and 1970s frequently uses the phrase
``cultural fairness'' when referring to parity between blacks and whites.
Other than Thomas \cite{thomas1973overprediction}, the test fairness 
literature of the 1960s and 1970s was typically concerned with race
rather than gender (although received attention later, e.g., \cite{willingham2013gender}). 
The role of culture in gender identity and gender presentation
has seen less consideration in ML fairness, but gender
labels raise ethical concerns \cite{hoffmann2017data, hamidi2018gender}.

Comparable to modern sentiment in the difficulties of measuring fairness, earlier decisions in the courtroom highlighted the impossibility of properly accounting for all factors that influence inequalities.  For example, in 1964, Illinois Fair Employment Practices Commission (FEPC) examiner found that Motorola had discriminated against Leon Myart, a black American, 
in his application to work at Motorola as an ``analyzer and phaser''. The examiner found that the 5 minute screening test that Myart took did not
account for inequalities and environmental factors of culturally deprived groups. The case was appealed to the Illinois Supreme Court, which found that
Myart actually passed the test, and so declined
to rule on the fairness of the test \cite{ash1966implications}.

\section{Fairness Gaps}\label{sec:gaps}

\subsection{Fairness and Unfairness} 
In mapping out earlier fairness approaches and their relationship to ML fairness, some conceptual gaps emerge.  One noticeable gap relates to the difference in framing between {\em fairness} and {\em unfairness}.  In earlier work on test fairness, there was a focus on defining measurements in terms of {\em unfair discrimination} and {\em unfair bias}, which brought with it the problem of uncovering sources of bias \cite{cleary1968investigation}.  In the 1970s, this developed into framings in terms of {\em fairness}, and the introduction of fairness criteria similar or identical to ML fairness criteria known today. However, returning to the idea of unfairness suggests several new areas of inquiry, including quantifying different kinds of unfairness and bias (such as content bias, selection system bias, etc., cf.~\cite{jencks1998racial}), and a shift in focus from outcomes to inputs and processes \cite{cojuharenco2013workplace}.  Quantifying types of {\em unfairness} may not only add to the problems that machine learning can address, but also accords with realities of sentencing and policing behind much of the fairness research today:  Individuals seeking justice do so when they believe that something has been {\em unfair}.  

\subsection{Differential Item Functioning} 

Another gap that becomes clear from the historical perspective is the lack of an analog to Differential Item Functioning (Section \ref{sec:60s_70s_diff}) in current ML fairness research. 
DIF was used by education professionals as a motivation for investigating causes of bias, and a modern-day analog might include unfairness interpretability in ML models.  An direct analog in ML could be to compare $P(X_i | R=r, A=a)$ for different
input features $X_i$, model outputs $R$ and subgroups $A$. For example, when predicting loan
repayment, this might involve comparing how income levels differ across subgroups 
for a given predicted likelihood of repaying the loan.

\subsection{Target Variable / Model Score Relationship} 

Another gap is the ways in which
the model (test) score and the target variable are related to each other. In many cases in ML fairness  and test fairness,
there are correspondences between pairs of criteria which differ
only in the roles played by the model (test) score $R$ and the target variable $Y$.
That is, one criterion can be transformed into another by swapping
the symbols $R$ and $Y$; for example, {\em separation} can be transformed into {\em sufficiency}: $A\perp R|Y \longrightarrow A\perp Y |R$. In this section we will refer to this type of correspondence as {\em ``converse''}, i.e.,
separation is the converse of sufficiency.

When viewed in this light, some asymmetries stand out:

\begin{itemize}
\item {\em Converse Cleary criterion}: Cleary's criterion considers the case
of a regression model that predicts a target variable $Y$ given test score $R$.
One could also consider the converse regression model (mentioned in passing by \cite{thorndike1971concepts}), which predicts model score
$R$ from ground truth $Y$, as an instrument for detecting bias.\footnote{The Cleary regression model and its converse are distinct except in the special
case where the magnitudes of the variables have been standardized.} The converse
Cleary condition would say that a test has connotations of {\em unfair}
for a subgroup if the converse
regression line has positive errors, i.e., for each given level of ground truth
ability, the test score is higher than the converse regression line predicts.
\item {\em Converse calibration}: In a regression scenario, the calibration condition
$P(Y=1|R=r, A=a)=r$ can be rewritten as $E(Y|R=r, A=a)=r$, or
$E(Y-r|R=r, A=a)=0$. The converse calibration 
condition is therefore $E(R-y|Y=y, A=a)=0$ for all subgroups $A=a$. In other words,
for each subgroup and level of ground truth performance $Y=y$, the expected
error in $R$'s prediction of the value $y$ is zero.
\end{itemize}
We point out these overlooked concepts not to advocate for their use, but to 
map out the geography of concepts related to fairness more completely.

\subsection{Compromises}

Darlington \cite{darlington1971another} points out that Thorndike's criterion is a
compromise between one criterion related
to sufficiency and one related to separation (see Section \ref{sec:70s} and Tables \ref{tab:fairness_criteria_summary} and \ref{tab:fairness_relationships}).
In general, a space of compromises is possible; in terms of
correlations, this might be modeled using a parameter $\lambda$: \vspace{-.5em}
\begin{equation}
\rho_{AR}=\rho_{AY}.\rho_{RY}^\lambda
\end{equation}
where $\lambda$ values of -1, 0, and 1 imply Darlington's definitions (1), (2) and (3), respectively.

This also suggests exploring interpolations between the
contrasting sufficiency and separation criteria. For example,
one way of parameterizing their interpolation is
in terms of binary confusion matrix outcomes.
\vspace{-.75em}
\begin{definition}
{\em $(\lambda_1,\lambda_2)$-Thorndikian
fairness}:
A binary classifier satisfies {\em $(\lambda_1,\lambda_2)$-Thorndikian
fairness}
with respect to demographic variable $A$ if both \vspace{.5em}

\begin{enumerate}
    \item[a)] $\displaystyle\frac{TP+\lambda_1 FP}{TP + \lambda_2 FN}$ is constant for all values of $A$ \vspace{.5em}, and
    \item[b)] $\displaystyle\frac{TN+\lambda_1 FN}{TN + \lambda_2 FP}$ is constant for all values of $A$.
\end{enumerate}
\end{definition}
Note that (1, 0)-Thorndikian fairness is equivalent to sufficiency,
while (0, 1)-Thorndikian fairness is equivalent to separation.

Petersen and Novick \cite{petersen1976evaluation} showed that $(1, 1)$-Thorndikian fairness requires that either
a) for each subgroup, the positive class is predicted in proportion to 
its ground truth rate; or b) every subgroup has the same ground truth
rate of positives. We can also consider relaxations of {\em $(\lambda_1,\lambda_2)$-Thorndikian fairness} in which only one of the
two conditions (a) or (b) is required to hold. For example, only requiring
condition (a) gives us a way of parameterizing compromises between equality 
of opportunity and predictive parity.

Our goal here is not to advocate for this particular model of compromise
between separation and sufficiency. Rather, since separation and sufficiency criteria can encode competing interests of different parties, our goal is to suggest that ML fairness
consider how to encode notions of compromise, which in some scenarios
might relate to the public's notion of fairness. We propose
that the economics literature on fair division might provide some useful ideas,
as has also been suggested by \cite{zafar2017parity}.
However, we do heed Darlington's \cite{darlington1971another}
warning that ``a compromise may end up satisfying nobody; psychometricians are not in the habit of agreeing on important definitions or theorems by compromise.'' This statement may be equally true of ML practitioners.

\section{Discussion}

This short review of historical connections in fairness suggest several concrete steps forward for future research in ML fairness:

\begin{enumerate}

    \item Developing methods to {\em explain} and {\em reduce}
    model unfairness 
    by focusing on the causes of {\em unfairness}.  To paraphrase Darlington's \cite{darlington1971another} question:
    ``What can be said about models that discriminate among
cultures at various levels?'' yields more actionable insights than ``What is a fair model?''
    This is related to research on causality in ML Fairness (see Section \ref{sec:equivalent_notions}),  but 
     including examination of full causal pathways, and processes that interact well before decision time. In other words: What causes the disparities?
    \item Drawing from earlier insights of Guion \cite{guion1966employment}, Thorndike \cite{thorndike1971concepts}, Cole \cite{cole1973bias}, Linn \cite{linn1973fair}, Jones \cite{jones1973moderated}, and Peterson \& Novick \cite{petersen1976evaluation} to expand fairness criteria to include model {\em context} and {\em use}.
    \item Building from earlier insights of 1970s researchers \cite{darlington1971another,hunter1976critical,linn1976insearch}
    to incorporate quantitative factors for the balance between fairness goals and other
    goals, such as a value system or a system of ethics.  This will likely include clearly articulating assumptions and choices, as recently proposed in \cite{MitchellEtAl2018}. 
    \item Diving more deeply into the question of how subgroups are defined, suggested as early as 1966 \cite{guion1966employment}, including questioning whether subgroups should be treated as discrete categories at all, and how intersectionality can be modeled.  This might include, for
example, how to quantify fairness along one dimension (e.g., age) conditioned on another
dimension (e.g., skin tone), as recent work has begun to address \cite{KearnsEtAl2018,FouldsPan2018}.
\end{enumerate}

\section{Conclusions}

The spike in interest in test fairness in the 1960s arose during a time of social and political upheaval, with quantitative definitions catalyzed in part by  U.S.\ federal anti-discrimination legislation in the 
domains of education and employment.  The rise of interest in fairness today has corresponded with public interest in the use of machine learning in criminal sentencing and predictive policing, including discussions around {\sc compas} \cite{larson2016analyzed, dieterich2016compas, corbettdavies2016computer} and 
{PredPol} \cite{oneill2016weapons,ensign2017runaway}.  Each era gave rise to its own notions of fairness and relevant subgroups, with overlapping ideas that are similar or identical.  In the 1960s and 1970s, the fascination with determining fairness ultimately died out as the work became less tied to the practical needs of society, politics and the law, and more tied to unambiguously identifying fairness. 

We conclude by reflecting on what further lessons the history of test fairness may
have for the future of ML fairness. Careful attention should be paid
to legal and public concerns about fairness. The experiences of the test
fairness field suggest that in the coming years, courts may start ruling
on the fairness of ML models. If technical definitions of fairness
stray too far from the public's perceptions of fairness, then the political will
to use scientific contributions in advance of public policy may be difficult
to obtain.  Perhaps ML practitioners should cautiously take heed from
Cole and Zieky's \cite{cole2001new} portrayal of developments in their field:
\begin{quote}
\fontsize{8.5}{10}\selectfont
    Members of the public continue to see apparently inappropriate interpretations of test scores and misuses of test results. They see this area as a primary fairness concern. However, the measurement profession has struggled to understand the nature of its responsibility in this area, and has generally not acted strongly against
    instances of misuse, nor has it acted in concert to attack misuses.
\end{quote}

We welcome broader debate on fairness that includes both technical and cultural causes, how the context and use of ML models further influence potential unfairness, and the suitability of the variables used in fairness research for capturing systemic unfairness. 
We agree with Linn's
\cite{linn1976insearch} argument from 1976 that values encoded by technical definitions 
should be made explicit.
By concretely relating fairness debates to ethical theories
and value systems (as done by \cite{hunter1976critical,zwick2016philosophical}), we can make discussions 
more accessible to the general public and 
to researchers of other disciplines, as well as 
helping our own ML Fairness community to be more attuned to our own implicit cultural biases. 

\section{Acknowledgements}

Thank you to Moritz Hardt and Shira Mitchell for invaluable conversations and insight.

\bibliographystyle{ACM-Reference-Format}
\bibliography{references}

\section*{Appendix a: Additional definitions of test fairness}
This appendix provides some details of fairness definitions included in Table~\ref{tab:fairness_criteria_summary} that were not introduced in the text 
of Section~\ref{sec:history}.

\subsection*{Einhorn and Bass}

In  1971, Einhorn and Bass 
\cite{einhorn1971methodological} noted that even if Cleary's criterion is satisfied, different rates of false positives and false negatives may be achieved for different subgroups due to differences in standard errors of estimate for the two subgroups.
That is, differences in variability around the common line of regression leads to different false positive and false negative rates. To address this, they propose a criterion based on achieving  equal false discovery rate, or as they put it, ``designated risk'', at the decision boundary. That is, $Prob(Y>y* | R=r_a*, A=a)$ is constant for all subgroups $a$.

\subsection*{Darlington's ``culturally optimum''}

Darlington \cite{darlington1971another} proposes that the
subjective value that one places on test validity (related to accuracy) and diversity can be scenario-specific.
He proposes a technique for eliciting these  value judgements, leading to a variable $k$ which measures the amount of tradeoff
in validity that is acceptable to increase diversity. He proposes
that the ``culturally optimum'' test is one that maximizes 
$\rho_{X(Y-kC)}$.

\subsection*{Jones}

In 1973, Jones  \cite{jones1973moderated}
proposed a ``general standard'' of fairness that is related to Thorndike's (and hence also related quota-based
definitions of fairness). In Jones' criterion, candidates are ranked in descending order both by test score and by ground truth. If an equal proportion of candidates from the subgroup are present in the top $n\%$ of both ranked lists then the test is fair ``at position $n$''. 
Jones' ``general standard'' of fairness requires that this hold for all values of $n$. Jones assumes a regression model relating test scores to ground truth, and also defines a weaker ``mean-fair'' criterion for a subgroup that
``the group's average predicted score
equals its average performance score on the [ground truth].''
 
\end{document}